\documentclass{ecai2010}
\usepackage{times}
\usepackage{latexsym}
\usepackage{amssymb}
\usepackage{amsmath}
\usepackage{tikz}
\usepackage{helvet,graphicx,multicol}
\usepackage{subfigure}

\newcommand{\G}{\mathcal{G}}
\newcommand{\J}{\mathcal{J}}
\newcommand{\T}{\mathcal{T}}

\newcommand{\I}{\mathcal{I}}
\newcommand{\SG}{\mathcal{S}_{\G}}
\newcommand{\ST}{\mathcal{S}_{\T}}

\newcommand{\AD}{App}
\newcommand{\DD}{Dis}

\newcommand{\TVG}{\ensuremath{\G=(V,E,\T,\rho,\zeta)}}

\newtheorem{definition}{Definition}

\begin{document}

\title{Rooting opinions in the minds: a cognitive model and a formal account of opinions and their dynamics}


\author{Francesca Giardini 
\institute{Department of Cognitive Science, Central European University, Budapest, Hungary, email: GiardiniF@ceu.hu}, 
Walter Quattrociocchi \institute{University of Siena, Italy, email: walter.quattrociocchi@unisi.it}
\and
Rosaria Conte \institute{ISTC-CNR, Rome, Italy, email:rosaria.conte@istc.cnr.it} 
}

\maketitle

\begin{abstract}
The study of opinions, their formation and change, is one of the defining topics addressed by social psychology, but in recent years other disciplines, like computer science and complexity, have tried to deal with this issue. 
Despite the flourishing of different models and theories in both fields, several key questions still remain unanswered. The understanding of how opinions change and the way they are affected by social influence are challenging issues requiring a thorough analysis of opinion per se but also of the way in which they travel between agents' minds and are modulated by these exchanges. 
To account for the two-faceted nature of opinions, which are mental entities undergoing complex social processes, we outline a preliminary model in which a cognitive theory of opinions is put forward and it is paired with a formal description of them and of their spreading among minds. Furthermore, investigating social influence also implies the necessity to account for the way in which people change their minds, as a consequence of interacting with other people, and the need to explain the higher or lower persistence of such changes.
\end{abstract}

\section{Introduction}
\label{sec:1}
The studies about opinions, persuasion and social influence are foundational and pressing issues in social psychology; however, within this discipline, the dynamics of opinions at the level of population has been underestimated. There are also other disciplines that have shown a great interest regarding such an issue, ranging from political science (\cite{lippmann}) passing through socio-physics (\cite{Castellano2007}) up to complexity science (\cite{Lorenz07}). 
Understanding opinions, describing how they are generated and revised, and how fare opinions travel over the social space both as a consequence of social influence and as one of the main means through which social influence unfolds, is crucial for grasping a deeper understanding of human social cognition and behaviors. 

Investigating opinions requires to take into account two levels of explanation: the individual and the social level. Social psychology has been mainly interested in explaining this first level, trying to describe the complex interplay of affective, cognitive and behavioral aspects that make opinions emerge. On the other hand, scholars from computer science and physics have tried to explain how different opinions can coexist or how they are modified through social interactions, treating opinions as objects that are exchanged and revised according to certain mechanisms that are quite far from the reality of cognitive and social processes. 
In both cases there is a reductionist fallacy that works in apparently different ways but it affects both these approaches, leading them to treat opinions either as a set of unrelated specific elements or as a unidimensional object that has nothing in common with a cognitive representation. 

We claim that opinions are highly dynamical representations resulting from the interplay of different mental representations and affected by the mental states of other individuals in the same network. 
Aim of this work is to provide an interdisciplinary account to describe how social influence leads to opinion formation, evolution and change. 
Moving from a characterization of opinions as mental representations with specific features, we will try to model how opinions are generated within the agents' minds (micro-level) and how they spread within a network of agents (macro-level). When explaining the emergence of macro-social phenomena we need to know what happens at the micro-level, i.e. what drives human actions and decisions in order to understand how individuals' representations and behaviors can give rise to socially complex phenomena and how those affect agents' actions. Without explaining how opinions are formed and manipulated within the individuals' minds, it is very difficult to account for the way in which they change as an effect of social influence. Our aim is to understand whether and how heterogeneous agents, endowed with different beliefs and goals, may come to share a given viewpoint and what consequences this sharing has on agents' behaviors. We are interested in providing answers, at least partially, to the following questions: What is an opinion? What mechanisms lead people to change their opinions? How can individuals resist to changes? What are the mechanisms of influence acting within and between individual minds? How does social impact affect agents' elaboration of new or contrasting information?

As opinion is still a debated concept within several disciplines, either its conceptualization or formalization are hard tasks. In particular, the actual instruments -e.g. metrics, formalisms – does not allow for a tight definition accounting for a) the relationships between opinions and other epistemic representations and b) their dynamics both at social and individual level.
In this paper we approach a preliminary {\em formal} definition of opinions by means of {\em Time Varying Graphs} \cite{CFQS2010}- e.g. a new formalism aimed at characterizing dynamically evolving systems as shown in \cite{AQ2010a,AQ2010b}. 

In section 2, a brief review of the state of the art is provided to introduce the main theories of opinions developed in the field of social psychology and to discuss more recent advances in opinion dynamics. 
Section 3 is devoted to the description of our model, in which a definition of opinions as specific mental representations and cognitively founded hypotheses about their diffusion and change will be put forward. In section 4 a preliminary formal account of how opinions are generated and how they can change is provided. In section 5 some conclusions are drawn and future directions are suggested.

\section{State of the Art}
\label{sec:2}

Social psychologists have devoted much attention to the study of opinions' formation and spreading, but a comprehensive and definite model allowing for an operational and generative account is still missing. Providing a comprehensive review of social psychology literature is beyond the scope of this work, but in this section we will discuss some of the main theories in order to underline how partial is the picture of opinions emerging from these studies.

In general, opinions are treated as synonyms for different mental objects, as beliefs \cite{fishbeinAjzen}, or more frequently, attitudes. Opinions are often conceptualized as attitudes \cite{mcguire}, \cite{olsonZanna}, \cite{Price} or they are used as interchangeable terms that have in common the fact of being affected by social influence and persuasion \cite{Pettyetal}. Allport \cite{allport} recognizes the difference between attitudes and opinions but he nonetheless considers the measurement of opinions as one way of identifying the strength and value of personal attitudes. An alternative view contrasts the affective content of attitudes with the more cognitive quality of opinions that involve some kind of conscious judgements \cite{Fleming}. Crespi \cite{crespi} considers individual opinions as "judgemental outcomes of an individual's transactions with the surrounding world" (p.19), emphasizing the interplay between what he calls an attitudinal system and the external world characterized by the presence of other agents and different subjective perceptions. Opinions are the outcomes of a judging process but this does not mean that they are necessarily rational or reasoned, although Crespi recognizes that they need to be consistent with the individual's beliefs, values and affective states.
As other authors already pointed out \cite{Masonetal.}, many models of opinion and social influence do not provide careful definitions of what an opinion is and how it is affected by social influence. This happens to be true also for theories of persuasion, like the social impact theory \cite{latane81}, a static theory of how social processes operate at the level of the individual at a given point in time. Part of this theory has been developed usign computational modeling by Nowak, Szamrej and Latané \cite{Latane90}. In their model, individuals change their attitudes as a consequence of other individuals' influence. In parallel with the idea that social influence is proportional to a multiplicative function of the strength, immediacy, and number of sources in a social force field \cite{latane81}, \cite{lavineLatane} suggest that each attitude within a cognitive structure is jointly determined by the strength, immediacy, and number of linked attitudes as individuals seek harmony, balance, or consistency among them. Although very interesting, this account fails to distinguish between attitudes and beliefs and does not explain how inconsistencies can be resolved.
The effect of communication on opinion formation has been addressed by different disciplines from within the social and the computational sciences, as well as complex systems science (for a review on attitude change models, see \cite{Masonetal.}). One of the first works on this topic has focused on polarization, i.e. the concentration of opinions by means of interaction, as one main effect of the "social influence'' \cite{festinger50}, whereas the Social Impact Theory' \cite{Latane90} proposes a more dynamic account, in which the amount of influence depends on the distance, number, and strength (i.e., persuasiveness) of influence sources. As stated in (\cite{Castellano2007}), an important variable, poorly controlled in current studies, is structure topology. Interactions are invariably assumed as either all-to-all or based on a spatial regular location (lattice), while more realistic scenarios are ignored.

Turning our attention to complex systems science, one of the most popular model applied to the aggregation of opinions is the bounded confidence model, presented in \cite{amblard01}. Much like previous studies, in this work agents exchanging information are modeled as likely to adjust their opinions only if the preceding and the received information are close enough to each other. Such aspect is modeled by introducing a real number $\epsilon$, which stands for tolerance or uncertainty (\cite{Castellano2007}) such that an agent with opinion $x$ interacts only with agents whose opinions is in the interval $] x - \epsilon ,  x + \epsilon[$.

The model we present in this paper extends the bounded confidence model by providing a cognitively plausible definition of opinion as mental representations and identifying their constitutive elements and their relationships.

\subsection{Main Advances}

This work aims at outlining a non-reductionist cognitive model of opinions and their dynamics. 
Differently from the models reviewed above, we first provide a definition of opinions as mental representations presenting specific features that make their revision and updating more or less easy and enduring. 
Moreover, grounding opinions in the minds allow us to take into account not only direct processes of revision triggered by the comparison with others' different opinions, i.e. social influence, but also revisions based upon changing in other mental representations supporting that opinion. 

The computational model introduced in this paper is intended to provide a preliminary unifying framework to define opinions and to characterize their dynamics in an easy but non-reductionist approach. Opinions in several models of opinion dynamics are considered to change according to social influence, we try to outline what is social influence and the way the social network structure affects the agents' opinions.

\section{A Cognitive Theory of Opinions}
Opinions can be described as configurations of an individual's beliefs, values and feelings that can be conditionally activated. This means that, for instance, starting from my feeling of aversion toward mathematics and as a consequence of having met a rude friend of friends who happened to teach math at school, when asked about my opinion on the time kids should spend in studying mathematics, I can form or, better, activate an opinion according to which the less time they spend the better it is. Opinions stem from the conditional activation of different kinds of mental representations, that can have a propositional content or, as in the case of attitudes and feelings, they can be more evaluative.
However, there is a specific feature that distinguishes an opinion from other kinds of mental objects. An opinion is an epistemic representation, thus it is a belief in which the truth-value is deemed to be uncertain. Opinions refer to objects of the external world that can not be told to be either true or false. This impossibility to say whether the content of a representation is true or false is what makes a mental representation an opinion, as opposed to a piece of knowledge, for instance. 
This basic feature can be paired with the presence of an attitude, i.e. an evaluative component that specifies whether the individual likes or dislikes the topic. In general, attitudes are present when the topic is somehow involving for the subject, so he is positively or negatively inclined toward it.

When this is not the case, we have "factual opinions", like in the following example. If someone is required to say when Mozart died, he can know the correct answer or not, but this is not a moot point. On the contrary, the causes of Mozart's death are debatable because without knowing where he was buried it is impossible to analyze the bones and to ascertain what killed him. This means that we know that Mozart died in 1791 but there are contrasting opinions about the causes of his death,  and, even if there exist one true opinion, none can tell which is the truth.
On the other hand, when opinions involve also evaluative components or facts, the opinions result from the activation of a pattern of related representations like knowledge, other opinions, but also goals. This view allows us to describe opinions as non-static patterns of relationships in which different representations are linked through a variety of different linkages. This work is meant to address the origin and changing of opinions thanks to these inter-relationships.

An opinion is characterized by the three following features. First, the truth value can not be verified (or it is not relevant). In general, opinions are representations whose truth value can not be assessed through direct experience. The topic of the opinion can not be experienced and then it is impossible to say whether a given object is true or false. If I ask someone about his opinion on the military intervention in Afghanistan, he can not tell me that his opinion, whether positive or negative, is true, because it is not possible to test an alternative state of the world in which the intervention has not taken place and then asses which state was the best. Nonetheless, he can tell me that he has a strong opinion or that he is very confident in it because he has many supporting beliefs (e.g. Talibans' regime had to be fighted, civilians needed the intervention, the world is a safer place after the intervention, etc) and even some goals (for instance, feeling safer) related with that opinion. This is to say that the lack of an assessable truth value is totally independent from the confidence one has in his opinions. We can have strong or weak opinions, but our confidence does not depend on the fact that something is known to be true, given the impossibility to assess its truth-value.

The second feature is the degree of confidence which is a subjective measure of the strength of belief and it expresses the exent to which one's opinion is resistant to change. The degree of confidence depends on the number of supporting representations, and the higher this number the stronger an opinion will be. Castelfranchi, Poggi \cite{CastelfranchiPoggi} made a distinction between confidence coming from the source and confidence coming from the degree of compatibility that a given belief has with pre-existing beliefs. It is interesting to notice that representations do not need to be about the same topic or to belong to the same set to form a coherent network. If we take the Afghanistan example, we can easily imagine that a negative opinion about the military intervention could be supported by a general belief about the right of other countries to intervene in internal disputes or by negative evaluations about the US foreign policy, or even by knowledge about the roles played by URSS and US in Afghanistan during the Cold War. These beliefs are not exclusively related to the target opinion and they can have stronger or weaker connections with other opinions. The stronger the confidence in these beliefs and the higher their number, the stronger will be the confidence in that opinion.

Finally, the sharing of an opinion, i.e. the extent to which a given opinion is considered shared, is another crucial feature. The sharing may heavily affect the degree of confidence, making people feel more confident because many other individuals have the same opinion. The sharing is the outcome of a process of social influence, through which agents' opinion are circulated within the social space and they can become more or less shared. This dimension is crucial, but it is also true that it carachterize other social beliefs, like reputation.

It is worth noticing that there are other kinds of beliefs that are really close to opinions but, at a closer investigation, there are some important differences. Reputation can be one of these, because it is shared and it is also carachterized by a varying degree of confidence. But, unlikely opinions, reputation has a truth value because it refers to someone's behaviors or actions that were actually exhibited (or that were reported as such, but we do not want to address here the issue of lying) and reported to other people. Reality matters in reputation, whereas it is much less relevant in opinions, as witnessed also by the fact that reputation does not have to be convincing (i.e. supported by some reasoning or arguments), whereas opinions have.

\section{Toward a Formal Definition}

\subsection{Preliminaries}

\subsubsection{Time Varying Graphs}
The temporal aspects of our opinion model is based on Time-Varying Graphs (TVG) formalism, a generic mathematical framework \cite{CFQS2010} designed to deal with the temporal dimension of networked data and to express their dynamics from an {\em interaction-centric} point of view \cite{ACFQS2010a}.

Consider a set of entities $V$ (or {\em nodes}), a set of relations $E$ between these entities ({\em edges}), and an alphabet $L$ accounting for any property such that a relation could have ({\em label}); that is, $E \subseteq V \times V \times L$.  $L$ can contain multi-valued elements. 

The relations (interactions) among entities are assumed to take place over a time dimension (continuos or discrete) $\T$ the {\em lifetime} of the system which is generally a subset of $\mathbb{N}$ (discrete-time systems) or $\mathbb{R}$ (continuous-time systems).  The dynamics of the system can subsequently be described by a time-varying graph, or TVG, \TVG, where
\begin{itemize}
\item $\rho: E \times \T \rightarrow \{0,1\}$, called {\em presence function}, indicates whether a given edge or node is available at a given time.
\item $\zeta: E \times \T \rightarrow \mathbb{T}$, called {\em latency function}, indicates the time it takes to cross a given edge if starting at a given date (the latency of an edge could vary in time).
\end{itemize}

\subsubsection{The underlying graph}
\label{sec:underlying-graph}
Given a TVG \TVG, the graph $G=(V,E)$ is called {\em underlying} graph of $\G$. This static graph should be seen as a sort of {\em footprint} of $\G$, which flattens the time dimension and indicates only the pairs of nodes that have relations at some time in a given time interval $\T$. In most studies and applications, $G$ is assumed to be connected; in general, this is not necessarily the case. Note that the connectivity  of $G=(V,E)$ does not imply that $\G$ is connected at a given time instant; in fact, $\G$ could be disconnected at all times. The lack of relationship, with regards to connectivity, between $\G$ and its footprint $G$ is even stronger: the fact that  $G=(V,E)$ is connected does not even imply that $\G$ is ``connected over time''.

\subsubsection{Edge-centric evolution}
From an edge point of view (relationships within epistemic representations), the evolution derives from variations of the availability. TVG defines the {\em available dates} of an edge $e$, noted $\I(e)$, as the union of all dates at which the edge is available, that is, $\I(e)= \{t \in \T : \rho(e,t)=1\}$. 
Given a multi-interval of availability $\I(e)=\{[t_1,t_2)\cup[t_3,t_4)...\}$, the sequence of dates $t_1,t_3,...$ is called {\em appearance dates} of $e$, noted $\AD(e)$, and the sequence of dates $t_2, t_4,...$ is called {\em disappearance dates} of $e$, noted $\DD(e)$. Finally, the sequence $t_1, t_2, t_3,...$ is called {\em characteristic dates} of $e$, noted $\ST(e)$.

\subsubsection{Graph-centric evolution}

From a global standpoint, the evolution of the system can be derived by a sequence of (static) graphs $\SG=G_{1}, G_{2}..$ where every $G_i$ corresponds to a static {\em snapshot} of $\G$ such that $e\in E_{G_i} \iff \rho_{[t_i,t_i+1)}(e)=1$, with two possible meanings for the $t_i$s: either the sequence of $t_i$s is a discretization of time (for example $t_i=i$); or it corresponds to the set of particular dates when topological events occur in the graph, in which case this sequence is equal to $sort(\cup\{\ST(e): e \in E\})$. In the latter case, the sequence is called  {\em characteristic dates} of $\G$, and noted $\ST(\G)$.

\subsection{Modeling Epistemic Representations}

An {\em opinion} is an epistemic representation of a state of the world with respect to a given object $p$. It is defined on a three dimensional space defined by: a) the {\em objective truth value} $T_o$, a {\em subjective truth value}, namely $T_s$ and a {\em degree of confidence} $d_{c}$ with respect to the object $p$.

More formally we can state that:

\begin{definition}
an epistemic representation of a state of the world $m \in M$ is a quadruplet ${p,T_o, T_s, d_{c}}$ defined by
a preposition $p$ related to a given object $O$, and two variable $T_o$ and $T_s$ defined on $\mathbb{R} $. The $d_{c} \in \mathbb{R}$ respectively quantifying the ``real`` truth value of an information, namely the objective truth value, the perceived truth values, and the degree of confidence, with respect to the preposition $p$.  
\end{definition}

By varying the dimensions of the domain of $T_o$ and $T_s$, we can define a taxonomy of the epistemic representation of the world that can be summarised as follows:

\begin{definition}
An epistemic representation $m_k = \{p,T_o,T_s,d_{c}\}$ is {\em knowledge} when $T_o = T_s$.
\end{definition}

\begin{definition}
An epistemic representation $m_b = \{p,T_o,T_s,d_{c}\}$ is a {\em belief} when $0 < T_o < 1 
\wedge 0 \leq T_s \leq 1$ .
\end{definition}

\begin{definition}
An epistemic representation $m_o = \{p,T_o,T_s,d_{c}\}$ is an \textbf{opinion} when $0 \leq T_o < 1 \wedge 0 \leq T_s \leq 1$.
\end{definition}

\subsection{Opinions and Individuals}

We can define an epistemic representation graph as a network of epistemic representation immerged in a dynamic network in a given time interval and the links state the correlation among them.
Let us consider  a  set  $V$ of mental representation (or nodes), interacting with one another over time.
Each {\em relation} among the mental representation can be formalized by a quadruplet $c=\{u,v,t_1,t_2\}$, where $u$ and $v$ are the involved mental representations (either beliefs, or knowledge or an opinion), $t_1$ is the time at which the correlation occurs, and $t_2$ the time at which the relation terminates.  A given pair of nodes can naturally be subject to several such interactions over time (and for generality, we allow these interactions to overlap). Given a time interval $\T=[t_a, t_b)\subseteq {\cal T}$ (where $t_a$ and $t_b$ may be either two dates, or one date and one infinity, or both infinities), the set $C(\T)$ (or simply  $C$) of all interactions  occurring during that time interval defines a set of intermittently-available edges $E(\T) \subseteq V\times V$, such that:

\begin{equation}
\begin{split}
  \forall u,v \in V, (u,v) \in E(\T) \\
\iff \exists t' \in [t_a,t_b), (u,v,t_1,t_2)\in C(\T) \ : \ t_1\le t'< t_2
\end{split}
\end{equation}

 \noindent that is, an edge $(u,v)$ exists iff at least one interaction between $u$ and $v$ occurs, or terminates, between $t_a$  and $t_b$. The intermittent availability of an edge $e=(u,v)\in E(\T)$ is described by the {\em presence function} $\rho: E(\T)\times \T \rightarrow \{0,1\}$ such that
  $ \forall t \in \T, e\in E(\T)$:

\begin{equation}
 \rho(e, t)=1 \iff    \exists  (u,v,t_1,t_2)\in C :  t_1\le t<t_2
\end{equation}

The triplet $\G=(V,E,\rho)$ is called an {\em epistemic representation graph}, and
the temporal domain  $\T=[t_a, t_b)$ of the function $\rho$, is  the {\em lifetime} of $\G$. 
We denote by $\G_{[t,t')}$  
the {\em mental representation subgraph} of $\G$ covering the period $[t_a,t_b) \cap [t,t')$

Hence, a sequence of couples $\J=\{(e_1,t_1), (e_2, t_2),...\}$, with $e_i \in E$ and $t_i \in \T$ for all $i$, is called a {\em journey} in $\G$ iff $\{e_1, e_2,...\}$ is a walk in $G$ and for all $i$, $\rho(e_i, t_i)=1$ and $t_{i+1} \ge t_{i}$. Journeys can be thought of as {\em paths over time} from a source node to a destination node (if the journey is finite).

Let us denote by $\J^*_\G$ the set of all possible journeys in an epistemic representation system $\G$. We will say that $\G$ {\em admits} a journey from a node $u$ to a node $v$, and note $\exists \J_{(u,v)} \in \J^*_\G$, if there exists at least one possible journey from $u$ to $v$ in $\G$.

\subsection{Opinion Dynamics and Society}

One of the most famous formalisms aimed at describing the process of persuasion is the ``Bounded Confidence Model'' (BCM) where agents exchanging information are modeled as likely to adjust their opinions only if the preceding and the received information are close enough to each other. Such an aspect is modeled by introducing a real number $\epsilon$ , which stands for tolerance or uncertainty  such that an agent with opinion $x$ interacts only with agents whose opinions is in the interval ]x − $\epsilon$ , x + $\epsilon$ [. 
Neverthless the wide, massive and cross-disciplinary use of the BCM (\cite{Lorenz07,Hu09}) ranging from ``viral marketing'' to to the Italians' opinions distortion played by controlled mass media (\cite{quattrociocchi2010e,brunetti2010,brunetti2010a,Hu09}). Such a model does not provide an explanation of the phenomena yielding to the tolerance value, it is just assumed as a static value.

In this work we will outline which are the factors affecting the acceptance or the refuse of one another opinion. 
In particular, how can we formalize comparison of two or more opinions? Recalling that a mental representation is a preposition with the truth value defined by two variable $T_o,T_s \in \mathbb{R} $ and $d_{c} \in \mathbb{R}$ respectively quantifying the ``real'' and the perceived truth value and the degree of confidence with respect to a given object or proposition.
And considering that such mental representations are modeled as set of time connected entities of the form $\G=(V,E,\rho)$ we can now provide some definitions aimed at describing the process of persuasion.

Assuming that an epistemic representation system, which is by nature adaptive, when facing with external events, reacts to the stimulus by activating only a subset of its components. For instance, consider the example where an agent $x$ is questioned by an agent $y$ about his opinion on a given target.

What does happen in the $x$'s mental representation system? How can we quantify $x$'s attitudes to change or not is opinions regarding a given matter of fact?

According to our model the epistemic representation system of $x$, as reaction to the external stimulus posed by the $y$'s question, will perform $journey$ within the elements that in its mind are related with the target of the question and on this base will be able to compare its opinion with the one owned by $y$.

\begin{definition}
{\em (relational-)connected component induced by an external event} in $\G_{x}$ is defined as a set of nodes $V'\subseteq V$ such that $\forall u,v \in V', \exists \J_{(u,v)} \in \J^*_\G$. Then $\G$ is said connected if it is itself a connected component ($V'=V$). 
\end{definition}

Since all nodes in $V'$ are defined by an objective truth value $T$ and a degree of confidence (perceived truth value) $d_g$ it is obvious that the resistence to an opinion to change is denoted by these values in all the nodes in $V'$.

\section{Conclusions}
\label{sec:5}

In this preliminary work we tried to sketch a cognitively grounded dynamic model of opinions, in which we defined these mental representations as carachterized by the presence of three specific features. Differently than psychological theories of opinions that usually provide rich definitions that are too complex to be reduced to measurable variables, we isolated three main constitutive elements that characterize this kind of mental representations. On the other hand, we tried to overcome the reductionist approach of opinion dynamic models, in which the richness of human cognitive processes is substituted by easy-to-compute factors poorly related to actual human behaviors. For this reason, we proposed to apply time-varying-graph to develop a formal model able to account for the way in which opinions are generated and change as a function of the presence and opinions of other agents in the network.

We are perfectly aware of the complexity of this issue and this work represents a preliminary attempt to merge the cognitive complexity of opinions with a rigorous formal approach, but there are many problems that we need to address. First, the cognitive model should be refined and specific hypotheses about opinion revision and diffusion should be put forward. Moreover, the robustness of the formal model will be tested and such a model will be implemented in cognitive multi-agent system in order to explore the parameter space upon which our model has been defined. Our ultimate aim is to build up a simulation environment in which agents endowed with heterogeneous representations of the external world interact and this leads to the creation of new opinions, the disappearing of some of the previous ones and, in general, to different distributions of representations in the population.

\section{Acknowledgments}
We thank two anonymous referees for their useful comments and participants of the Workshop on Gossip and Opinions (Rome, June 2010, ISTC CNR). A particular thanks goes to Isabella Poggi and Cristiano Castelfranchi for their comments.

\bibliographystyle{plain}
\bibliography{biblio}

\begin{thebibliography}{10}

\bibitem{Masonetal.}
E.~Smith A.~Mason, F.Conrey.
\newblock Situating social influence processes: Dynamic, multidirectional flows
  of influence within social networks.
\newblock {\em Personality and Social Psychology Review}, 11(279-300), 2007.

\bibitem{Latane90}
B.~Latan{\'e} A.~Nowak, J.~Szamrej.
\newblock From private attitude to public opinion: A dynamic theory of social
  impact.
\newblock {\em Psychological Review}, 97:362--376, 1990.

\bibitem{allport}
G.W. Allport.
\newblock {\em Readings in attitude theory and measurement}, chapter Attitude,
  pages 1--13.
\newblock Wiley, 1967.

\bibitem{brunetti2010}
S.~Brunetti, E.~Lodi, and W.~Quattrociocchi.
\newblock Multicolored dynamos on toroidal meshes.
\newblock {\em CoRR}, abs/1012.4404, 2010.

\bibitem{brunetti2010a}
S.~Brunetti, E.~Lodi, and W.~Quattrociocchi.
\newblock Dynamic monopolies in colored tori.
\newblock {\em APDCM - Alaska}, 2011.

\bibitem{CastelfranchiPoggi}
I.~Poggi C.~Castelfranchi~C.
\newblock {\em Bugie, finzioni e sotterfugi. Per una scienza dell'inganno}.
\newblock Carocci Editore, 1998.

\bibitem{Castellano2007}
V.~Loreto C.~Castellano, S.~Fortunato.
\newblock {Statistical physics of social dynamics}.
\newblock {\em Reviews of Modern Physics}, 81(2):591+, June 2009.

\bibitem{CFQS2010}
A.~Casteigts, P.~Flocchini, W.~Quattrociocchi, and N.~Santoro.
\newblock Time-varying graphs and dynamic networks.
\newblock {\em Arxiv arXiv:1012.0009}, 2010.

\bibitem{crespi}
I.~Crespi.
\newblock {\em The public opinion process. How the people speak}.
\newblock Lawrence Erlbaum Associates, 1997.

\bibitem{amblard01}
G.~Deffuant, D.~Neau, F.~Amblard, and G.~Weisbuch.
\newblock Mixing beliefs among interacting agents.
\newblock {\em Advances in Complex Systems}, 3:87--98, 2001.

\bibitem{festinger50}
L.~Festinger, S.~Schachter, and K.~Back.
\newblock {\em Social Pressures in Informal Groups: A Study of Human Factors in
  Housing}.
\newblock Harper, New York, NY, USA, 1950.

\bibitem{Fleming}
D.~Fleming.
\newblock Attitude: The history of a concept.
\newblock {\em Perspectives in American History}, 1:287--365, 1967.

\bibitem{lavineLatane}
B.~Latan{\'e} H.~Lavine.
\newblock A cognitive-social theory of public opinion: Dynamic impact and
  cognitive structure.
\newblock {\em Journal of Communication}, 46:48--56, 1996.

\bibitem{Hu09}
H.~Hu and X.~Wang.
\newblock {Discrete opinion dynamics on networks based on social influence}.
\newblock {\em Journal of Physics A: Mathematical and Theoretical},
  42(22):225005+, June 2009.

\bibitem{olsonZanna}
M.P.~Zanna J.M.~Olson.
\newblock Attitudes and attitude change.
\newblock {\em Annual Review of Psychology}, 44:117--154, 1993.

\bibitem{latane81}
B.~Latan{\'e}.
\newblock The psychology of social impact.
\newblock {\em American Psychologist}, 36:343--356, 1981.

\bibitem{lippmann}
W.~Lippmann.
\newblock {\em Public opinion}.
\newblock Penguin Books, 1946.

\bibitem{Lorenz07}
J.~Lorenz.
\newblock {Continuous opinion dynamics of multidimensional allocation problems
  under bounded confidence: More dimensions lead to better chances for
  consensus}.
\newblock Aug 2007.

\bibitem{mcguire}
W.~McGuire.
\newblock The vicissitudes of attitudes and similar representational constructs
  in twentieth century psychology.
\newblock {\em European Journal of Social Psychology}, 16:89--139, 1986.

\bibitem{fishbeinAjzen}
I.~Ajzen M.Fishbein.
\newblock {\em Belief, Attitude, Intention, and Behavior: An Introduction to
  Theory and Research}.
\newblock Reading, MA: Addison-Wesley, 1975.

\bibitem{Price}
V.~Price.
\newblock {\em Communication concepts 4: Public opinion}.
\newblock Sage, 1992.

\bibitem{AQ2010b}
W.~Quattrociocchi and F.~Amblard.
\newblock Emergence through selection: The evolution of a scientific challenge.
\newblock {\em Arxiv arXiv:1102.0257}, Dec 2010.

\bibitem{AQ2010a}
W.~Quattrociocchi and F.~Amblard.
\newblock Selection in scientific networks.
\newblock {\em Arxiv arXiv:1012.4396v1}, Dec 2010.

\bibitem{quattrociocchi2010e}
W.~Quattrociocchi, R.~Conte, and E.~Lodi.
\newblock Simulating opinion dynamics in heterogeneous communication systems.
\newblock {\em ECCS 2010 - Lisbon Portugal}, 2010.

\bibitem{Pettyetal}
L.R.~Fabrigar R.E.~Petty, D.T.~Wegener.
\newblock Attitudes and attitude change.
\newblock {\em Annual Review of Psychology}, 48:609--647, 1997.

\bibitem{ACFQS2010a}
N.~Santoro, W.~Quattrociocchi, P.~Flocchini, A.~Casteigts, and F.~Amblard.
\newblock Time varying graphs and social network analysis: Temporal indicators
  and metrics.
\newblock {\em Arxiv arXiv:1102.0629}, 2010.

\end{thebibliography}
\end{document}